\documentclass[submission,copyright,creativecommons]{eptcs}
\providecommand{\event}{ICLP DC 2019} 
\usepackage{underscore}           
\usepackage{graphicx}
\usepackage{amsmath}
\usepackage{hyperref}

\newcommand{\imgpad}[0]{\vspace{2mm}}

\title{Experimenting with\\ Constraint Programming on GPU}
\author{Fabio Tardivo
	\institute{Department of Computer Science\\
		New Mexico State University\\
		Las Cruces, United States}
	\email{ftardivo@nmsu.edu}}
\def\titlerunning{Experimenting with Constraint Programming on GPU}
\def\authorrunning{Fabio Tardivo}
\begin{document}
	\maketitle
	
	\begin{abstract}
	The focus of my PhD thesis is on exploring parallel approaches to efficiently solve problems modeled by constraints and presenting a new proposal. Current solvers are very advanced; they are carefully designed to effectively manage the high-level problems' description and include refined strategies to avoid useless work. Despite this, finding a solution can take an unacceptable amount of time. Parallelization can mitigate this problem when the instance of the problem modeled is large, as it happens in real world problems. It is done by propagating constraints in parallel and concurrently exploring different parts of the search space.	I am developing on a constraint solver that exploits the many cores available on Graphics Processing Units (GPU) to speed up the search.
	\end{abstract}
	
\section{Introduction and Background}
\subsection{Constraint programming}
Constraint programming is a declarative programming paradigm, focused on problem modeling, instead of specifying the steps to find the solution.

A Constraint Satisfaction Problem (CSP) is a triple $ P = (\mathcal{X},\mathcal{D},\mathcal{C})$
where $\mathcal{X}$ is a set of variables, $\mathcal{D}$ is the set of the variables' domains and $\mathcal{C} $ is a set of constraints. Formally a constraint specifies a subset of the cartesian product of the involved domains:
	\begin{align*}
		\mathcal{X} &= \{x_1, x_2\} \\
		\mathcal{D} &= \{d_1, d_2\} = \{\{1, 2, 3, 4, 5\}, \{1, 2, 3, 4, 5, 6, 7, 8, 9, 10\}\}\\
		\mathcal{C} &= \{c_1, c_2\} = \{x_1 > 3, x_1 < x_2\} = \{ \{4, 5\}, \{(1,2), (1,3),\dots , (5,10)\}\}
	\end{align*} 
A solution is an assignment of elements of their domains to variables that satisfy all the constraints.

A constraint solver is a procedure capable of returning the solution(s) to a CSP. It is build on three main operations:
\begin{description}
	\item[Domains reduction] The solver non-deterministically assigns a variable with a value in its domain.	
	\item[Constraints propagation] The solver checks and possibly removes values that cannot occur in any solution.
	\item[Backtracking] The solver restores the domains in case of a dead-end and records the solution in case of a success.
\end{description}

\subsection{CUDA}
The increasing performance ratio between GPU and CPU encouraged researchers to use GPUs for non-graphics computations \cite{larsen2001fast}. Computed Unified Device Architecture (CUDA) is an API to perform general computing on Nvidia GPUs.\\
From the CUDA point of view, the GPU is made of SIMD-like processors, each with 32 cores:
\begin{center}
	\imgpad
	\includegraphics[width=0.5\linewidth]{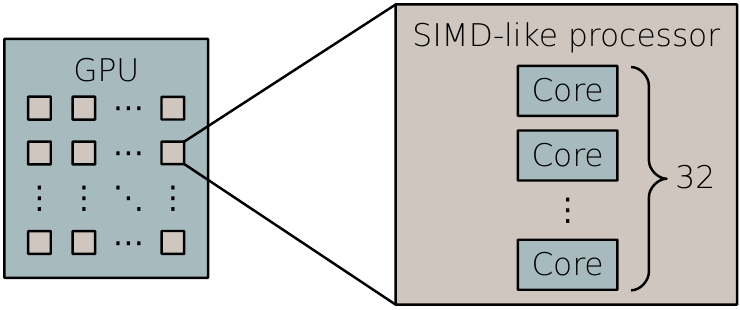}
	\imgpad
\end{center}	
The memory hierarchy is mainly composed of 3 layers: Shared Memory, L2 Cache and Global Memory:
\begin{center}
	\imgpad
	\includegraphics[width=0.5\linewidth]{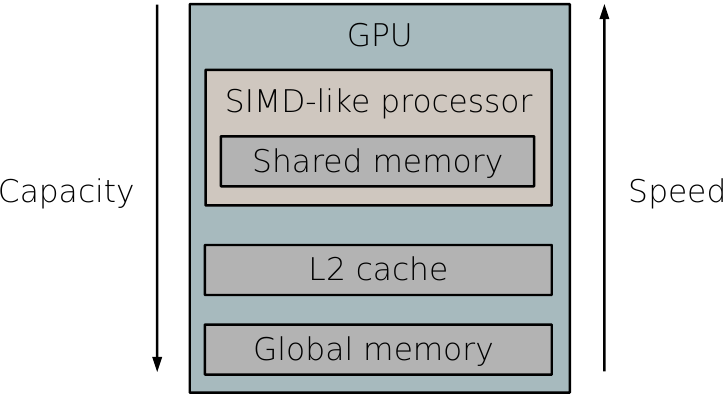}
	\imgpad
\end{center}

\subsection{Parallel Constraint Programming}
Over the years many approaches to parallel constraint solving have been explored. They are usually designed to run on computational clusters, and can be classified according to which part of the solving procedure is parallelized.
\begin{description}
	\item[Parallel  propagation] There are two main approaches to parallelizing the constraint propagation. The first is to partition the variables among the computational units, duplicate the constraints and communicate the removed values \cite{nguyen1998distributed}. The second is to partition the constraints among the computational nodes, duplicate variables  and communicate the removed values \cite{ruiz1998parallel}.	
	
	\item[Parallel search] This approach can be generalized as a search space partition, where each part is assigned to a worker that acts as a standard solver. To address unbalanced workloads, the authors use centralized task dispatch \cite{schulte2000parallel}, tasks pool \cite{michel2007parallelizing} and tasks with priorities \cite{chu2009confidence}.
	
	\item[Portfolio method] This method uses multiple solvers to find solutions. The solver configuration relay on a performance-problem database \cite{o2008using} or on the average performances with similar problems \cite{amadini2015sunny}.
	
	\pagebreak
	\item[GPU] This technique uses the GPU to accelerate the search process. The first implementation of this type of solver \cite{campeotto2014exploring} only uses the GPU to  propagate complex constraints. In subsequent applications, GPU has been used to perform local search heuristics like Adaptive Search \cite{arbelaez2014gpu} and Large Neighborhood Search \cite{campeotto2014gpu}. GPU was also used for SAT problem \cite{journals/jetai/PaluDFP15}, in detail the GPU was used to parallelize the unit-propagation and the exploration of the  search tree's low part.
	
\end{description}
For a more complete and detailed survey, the reader can refer to \cite{pcp,DBLP:books/sp/18/DovierFP18,DBLP:books/sp/HS2018}.

\section{CUBICS}

In this section the solver named CUda BasIc Constraint Solver (CUBICS) is described. The first part gives an high level overview of the solver's workflow, the second part describes how constraint propagation and backtracking are parallelized and the last part is about some low level implementation details.

\subsection{High-level design}
The constraint solver performs the entire computation on GPU, since CPU-GPU data transfer can nullify the time gained by the parallelization.
First, the CPU parses the problems, creates the data structures and moves them to the GPU's memory.\\
Control is then moved to the GPU, which performs labeling, constraint propagation, and backtracking. Here, constraint propagation is fully parallelized, while backtracking is partially parallelized and labeling is not parallelized at all, due to its sequential nature. \\
Once a solution is found, the control moves back to CPU which prints the result and potentially launches a new search. 
\begin{center}
	\imgpad
	\includegraphics[width=0.60\linewidth]{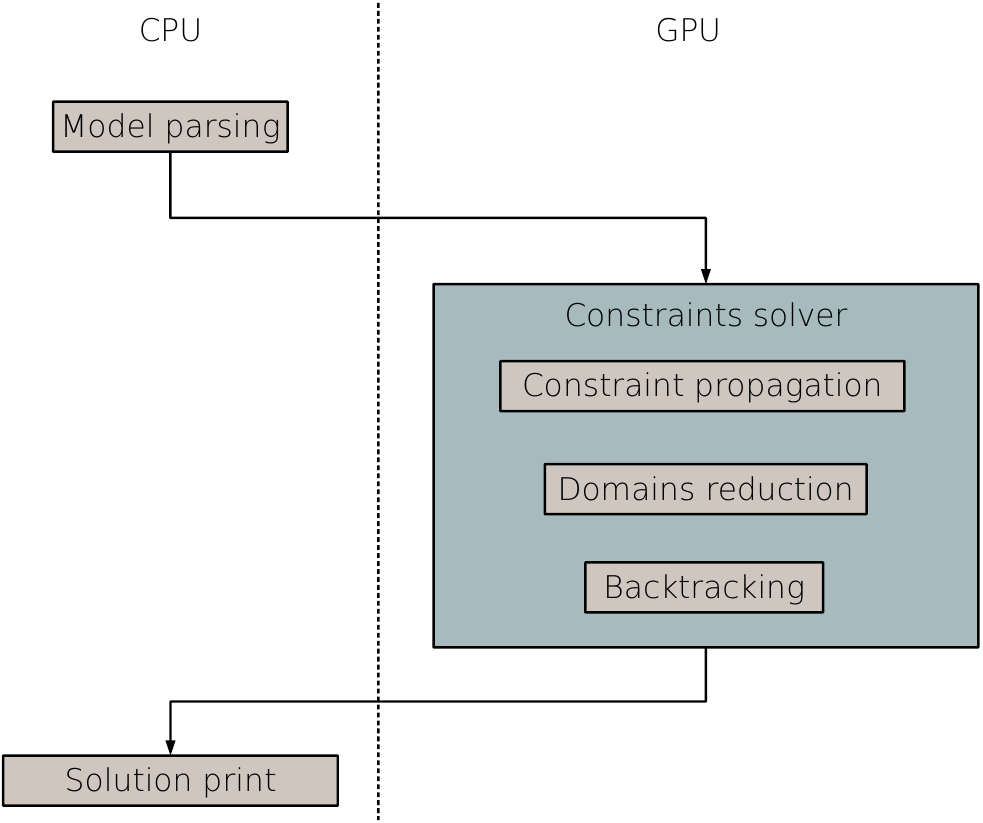}
	\imgpad
\end{center}

\pagebreak
\subsection{Parallelization}
CUDA 5.0 introduced a new feature, referred to as Dynamic Parallelism. This feature allows a GPU thread to launch other GPU threads without returning the control to the CPU. \\
This mechanism is mainly exploited in CUBICS. Instead of propagating each constraint sequentially, the GPU propagates all of the constraints in parallel. The main GPU thread spawns other threads so that constraints of the same type are propagated by the same GPU processor. This is because  each GPU processor behaves in a SIMD-like way, so it is desirable that each thread follows the same execution path.
\begin{center}
	\imgpad
	\includegraphics[width=0.6\linewidth]{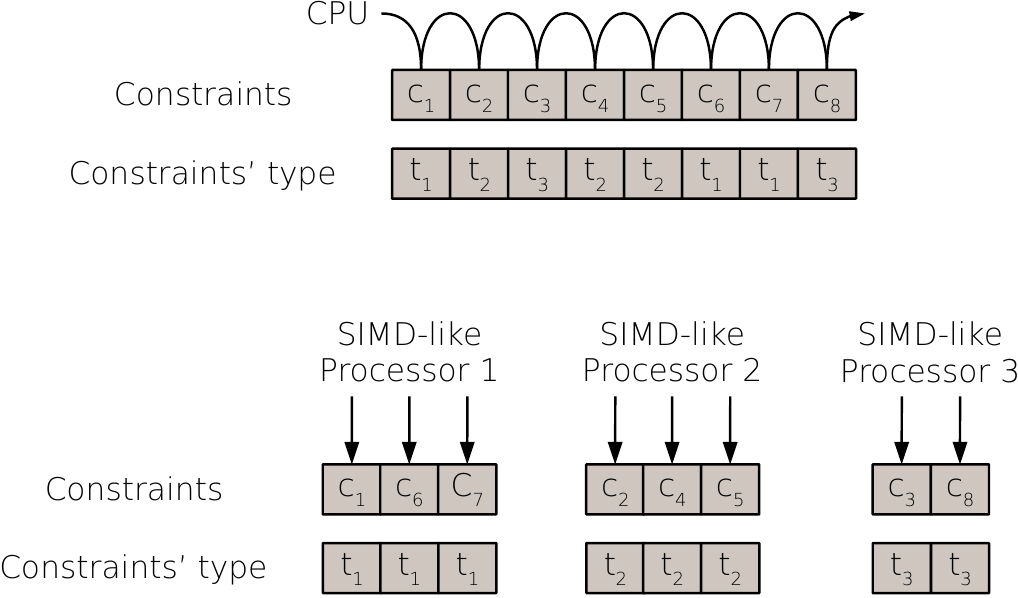}
	\imgpad
\end{center}
Dynamic Parallelism is also used in backtracking to save and restore domains. This is done by moving from a single global stack to multiple stacks, one for each domain. In such configuration, the main GPU thread spawns as many threads as the domains to manage, without regard to the GPU processor they will be running on, since the code is common.
\begin{center}
	\imgpad
	\includegraphics[width=0.8\linewidth]{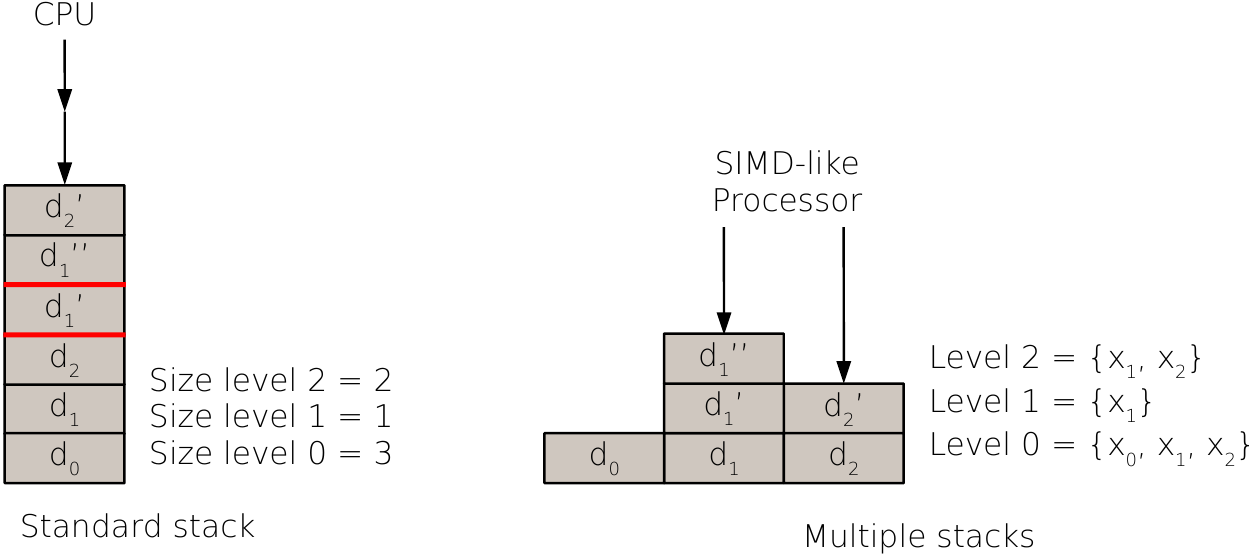}
	\imgpad
\end{center}

\subsection{Implementation details}
CUDA 6.0 introduced Unified Memory, which is a mechanism that abstracts the CPU and GPU memory, providing a unique memory address space. The data are initially stored in the CPU memory and implicitly moved on demand between the CPU and the GPU by the CUDA runtime system. This mechanism is used to facilitate both the creation of data structures and solutions retrieval.\\
One of the fundamental features of GPUs is the shared memory. It is an extremely fast on-chip memory, freely manageable by the programmer. This memory is exploited by the constraints propagation to cache the most accessed values. In case a constraint propagation needs to cache a significant amount of data, the solver reduces the number of threads per processor so the memory can accommodate the data.

\section{Future works}
There are many ways to extend the solver but few of them properly exploit the GPU architecture.\\
The first direction we will explore is the parallelization of the Large Neighborhood Search (LNS) methodology \cite{dekker2018solver}. Once the solver finds a solution, it randomly resets some domains and performs another search looking for better solutions. This process can be parallelized creating many copies of the CSP, each with different reset domains.
This approach explores much more search space at the same time, and allows for the sharing of the intermediate solutions so as to avoid looking for worse solutions, which increase the overall quality of the solutions. LNS has been successfully implemented on GPU in \cite{campeotto2014gpu}.\\
Another point that looks promising involves how the constraints are propagated. There are some global constraints for which propagation complexity is exponential, so the propagation algorithms for such constraints are polynomial approximations of the originals. With GPU, it will be possible to use more sophisticated algorithms that remove more values from the domains, improving the overall search time.\\

\bibliographystyle{eptcs}
\bibliography{references}

\end{document}